\documentclass[10pt,twocolumn,letterpaper]{article}

\usepackage{cvpr}              
\usepackage{adjustbox}
\usepackage{multirow, makecell}
\usepackage{pifont}
\usepackage{algorithm}
\usepackage{algorithmic}
\usepackage{enumitem}
\usepackage{amsmath,amssymb,mathtools}
\usepackage[table]{xcolor}
\usepackage{url}
\definecolor{cvprblue}{rgb}{0.21,0.49,0.74}
\usepackage[pagebackref,breaklinks,colorlinks,allcolors=cvprblue]{hyperref}

\def\paperID{43828} 
\def\confName{CVPR}
\def\confYear{2026}

\title{Retrieval-Driven Training-Free AI-Generated Video Attribution}

\author{
	Renxi Cheng\textsuperscript{1}, Chaolei Han\textsuperscript{1}, Jie Gui\textsuperscript{1,2,3}, Hongsong Wang\textsuperscript{4,5}\\
    $^{1}$School of Cyber Science and Engineering, Southeast University, Nanjing 210096, China\\
    $^{2}$Purple Mountain Laboratories, Nanjing 210000, China \\
	$^{3}$Engineering Research Center of Blockchain Application, Supervision And Management\\ (Southeast University), Ministry of Education, China \\
	$^{4}$School of Computer Science and Engineering, Southeast University, Nanjing 210096, China \\
	$^{5}$Key Laboratory of New Generation Artificial Intelligence Technology and Its Interdisciplinary \\
	Applications (Southeast University), Ministry of Education, China \\ 
	\tt\small\{renxi, chaoleihan, guijie, hongsongwang\}@seu.edu.cn \\
}

\begin{document}
\maketitle

\begin{abstract}
AI-generated videos are becoming increasingly realistic and difficult to distinguish from authentic ones, which facilitates malicious misuse and poses growing threats to cybersecurity and social governance. Attributing AI-generated videos to their specific generative sources is therefore of critical importance for forensic investigation and legal regulation. However, most existing visual attribution methods focus on images and particularly rely on the image generation model, thereby lacking the ability to generalize to large-scale AI-generated video data. To address these limitations, we introduce an training-free AI-generated video attribution paradigm. Specifically, we formulates AI-generated video attribution as an instance retrieval task, and design a generative fingerprint-based pipeline. This pipeline consists of an adapted orthogonal color transformation, multi-scale quantized residual generation, and temporal-semantic aggregation, progressively capturing and integrating artifacts introduced by generative models across video frames. Extensive experiments on the GenVidBench benchmark demonstrate that our method achieves strong performance in both AI-generated video detection and attribution, outperforming existing state-of-the-art methods with a Rank-1 accuracy of 20.5$\%$ and a mean Average Precision of 16.6$\%$. The code is at \url{https://github.com/renxi-seu/Video_Attribution}.
\end{abstract}

\section{Introduction}

With the rapid evolution of video generative models~\cite{morphstudio,xing2024dynamicrafter,chen2023seine,StepVideo,Animatediff,OpenSora,LTX,HYV,wan2.1}, AI-generated videos are approaching real videos in both visual quality and perceptual realism, making low-cost video synthesis increasingly accessible. However, this also makes forged content more prone to misuse in disinformation, identity impersonation, and malicious manipulation~\cite{misuse-1,misuse-2,earthquake}, posing growing threats to cybersecurity and social governance. Therefore, there is an urgent need to develop effective defensive mechanisms to mitigate these potential risks. 


\begin{figure}[t]
    \centering
    \includegraphics[width=\linewidth]{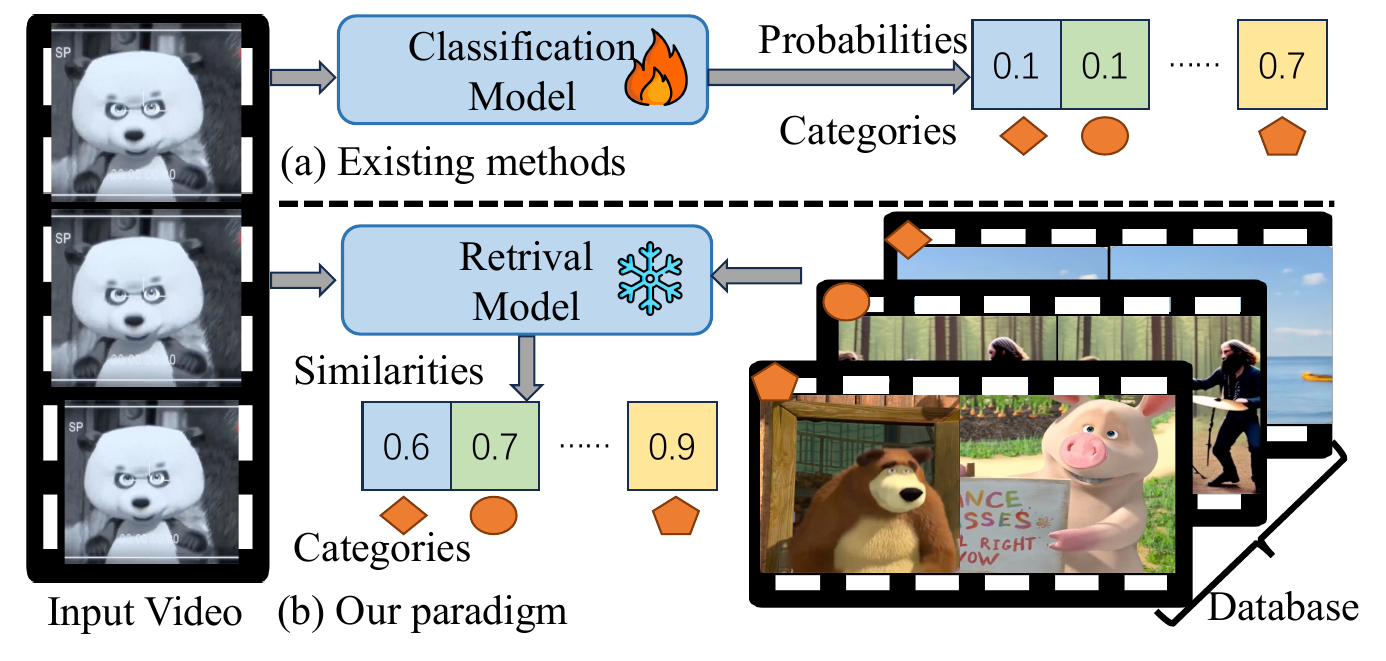}
    \caption{Comparison between existing AI-generated video attribution methods and our training-free retrieval-based framework. Our method is highly scalable, as new generators can be accommodated by simply registering reference samples without model retraining. }
    \label{fig:intro}
\end{figure}

As a fundamental and easily deployable passive defense, AI-generated video detection receives extensive attention. 
Existing methods mainly focus on AI-generated image detection~\cite{MING2026113795,CHEN2026112755}, or exploit video artifacts from spatial~\cite{li2018exposing,rossler2019faceforensics++,frank2020leveraging,LIU2026113813} and temporal~\cite{ma2025detectingaigeneratedvideoframe,ji2024distinguishfakevideosunleashing,xue2025videoforgerydetectionoptical} perspectives, achieving strong performance in distinguishing real videos from generated ones. 
However, such methods remain limited to binary authenticity classification and fail to identify the underlying generative source, which constrains their practical value in forensic and regulatory scenarios. Consequently, AI-generated video attribution has become increasingly important; however, it remains underexplored so far.

A few existing video attribution methods~\cite{VGM,vahdati2024beyond,kundu2025sagasourceattributiongenerative} adopt a supervised classification paradigm, where models are trained to learn source-discriminative artifacts and assign each video to a predefined set of generative sources. While these approaches can achieve promising performance when candidate generators are known and sufficiently represented during training, they still suffer from two fundamental limitations.
First, they rely heavily on large-scale labeled datasets with source annotations, which are costly to collect and maintain as generation models evolve rapidly. 
Second, these models are restricted to closed-set scenarios and tend to overfit to a limited number of video generators.

Motivated by these limitations, we revisit the nature of the attribution task itself. 
We argue that \textit{attribution is more naturally formulated as a source matching problem than a closed-set classification problem: a query video should be attributed to the generative source whose fingerprint it most closely resembles. }
Under this view, if generator-specific artifacts can be effectively exposed, attribution can be reformulated as retrieval over a reference database, avoiding the need to learn source-specific classifiers.

Based on this insight, we propose a training-free retrieval-driven framework for AI-generated video attribution, as shown in Fig.~\ref{fig:intro}. 
Since retrieval performance critically depends on the quality of the representation, the central challenge is to construct video features that can reliably expose and preserve generator-specific fingerprints.
We observe that although modern video generative models achieve high semantic fidelity, they often fail to fully maintain the consistency of low-level statistical patterns during synthesis, leaving behind subtle yet stable artifacts. Such artifacts are closely related to the underlying generation mechanisms and therefore provide a natural basis for source attribution.
To capture these signals, we design a fingerprint-aware representation pipeline that progressively enhances and aggregates generator-specific artifacts from spatial and temporal perspectives. In particular, we first introduce an adapted orthogonal color transformation to expose informative channel-wise variations. We then perform multi-scale quantized residual generation to amplify fine-grained discrepancies that are difficult to observe in the original frame space. Finally, we employ temporal-semantic aggregation to integrate frame-level artifacts into a coherent video-level fingerprint representation. Based on the resulting representations, we leverage a pre-trained video encoder to extract feature embeddings and perform similarity-based retrieval over the reference database.
Moreover, beyond attribution, the proposed framework naturally supports AI-generated video detection, as real videos tend to remain dissimilar to the fingerprints of registered generators. Extensive experiments on GenVidBench demonstrate that our method achieves strong performance on both attribution and detection tasks.

Our contributions are summarized as follows:
\begin{itemize}
\item \textbf{Open-set paradigm for AI-generated video attribution:} We study AI-generated video attribution in an open-set scenario by formulating it as an instance retrieval task.
\item \textbf{Training-free generative fingerprint-based pipeline:} We present a training-free pipeline that leverages model-specific generative artifacts by projecting video frames into an adaptive color space for residual computation.
\item \textbf{Few-shot benchmarks for detection and attribution:} We establish benchmarks for few-shot AI-generated video detection and attribution, and our approach outperforms baselines by 20.5\% in Rank-1 accuracy and 16.6\% in mAP.
\end{itemize}

\section{Related Work}

\subsection{AI-Generated Video Detection}

Existing methods for AI-generated video detection initially focus on static spatial features, including traditional image processing techniques~\cite{li2018exposing}, CNN-based models~\cite{rossler2019faceforensics++}, and frequency-domain analysis~\cite{frank2020leveraging}.
To address the lack of temporal dynamics, researchers develop a pipeline to combine spatial and temporal information, including single branch serial structure and dual branch parallel structure. 
For single branch serial structure, Sabir~\textit{et al.}~\cite{sabir2019recurrent} pioneer the use of recurrent convolutional networks to exploit temporal discrepancies across frames. Similarly, Vahdati~\textit{et al.}~\cite{vahdati2024beyond} and Ma~\textit{et al.}~\cite{ma2025detectingaigeneratedvideoframe} focus on firstly extracting spatial artifacts and subsequently modeling their temporal evolution to detect inconsistencies. Other advancements like DeMamba~\cite{DeMamba} further refines this serial process by utilizing state space models for efficient long-range dependency modeling, and UNITE~\cite{kundu2025universalsyntheticvideodetector} employs uncertainty-aware representation learning to robustly integrate visual and temporal clues. Beyond standard convolution-based sequences, FAST~\cite{yu2021frequency} is the first transformer-based framework designed to simultaneously exploit spatial, temporal, and frequency-aware manipulation traces. For dual branch parallel structure, AIGVDet~\cite{bai2024aigeneratedvideodetectionspatiotemporal} employs two sub-detectors to independently identify anomalies in spatial and optical flow domains, while DuB3D~\cite{ji2024distinguishfakevideosunleashing} utilizes a dual-branch 3-D Transformer to adaptively fuse spatio-temporal data and dense optical flow. He~\textit{et al.}~\cite{he2024exposingaigeneratedvideosbenchmark} fuses local motion and global appearance to expose defects at different scales, which improves the performance and robustness. Xue~\textit{et al.}~\cite{xue2025videoforgerydetectionoptical} employ second-order optical flow residuals to amplify subtle, high-frequency motion anomalies, resulting in higher discriminative sensitivity across diverse text-to-video and image-to-video tasks.
However, these methods remain limited to distinguishing real from AI-generated videos, without identifying the specific generative source behind them. 

\begin{figure*}[t]
    \centering
    \includegraphics[width=\linewidth]{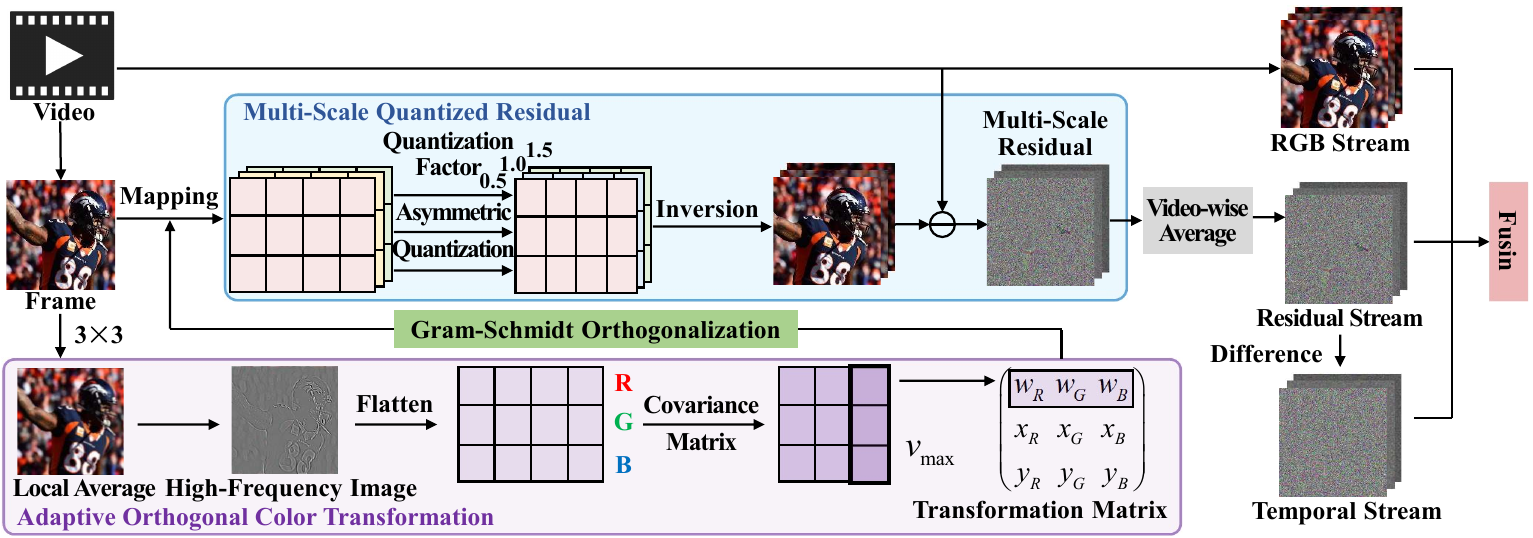}
    \caption{Overview of our fingerprint-aware representation pipeline. (1) The input video frames are first computed with an error map to generate the transformation matrix of the adaptive color space. (2) In the adaptive color space, multi-scale and channel-asymmetric quantization is performed on input video frames. Afterwards, the residual frames are generated. (3) RGB frames, residual frames, and a temporal stream derived from the residual frames are fused through weighted aggregation.}
    \label{fig:method}
\end{figure*}

\subsection{AI-Generated Visual Attribution}

The research of generative model attribution mainly focuses on the field of images. RONAN~\cite{wang2023did} proposes a universal attribution method that utilizes reverse-engineering and reconstruction loss to determine the source of images. This architecture solves the limitation of previous attribution methods that rely on specific generative models. In response to the problem that traditional attribution methods are limited by close-set scenarios and cannot recognize new models, many attribution works based on open-set scenarios begin to emerge. Girish~\textit{et al.}~\cite{girish2021towards} propose an iterative framework that successfully solves the identification of unknown source models in open-set scenarios. Yang~\textit{et al.}~\cite{yang2023progressive} further propose a progressive open-space expansion scheme to better distinguish known and unknown models in open-set scenarios. However, these methods cannot be directly extended to video attribution, as videos involve not only richer spatial structures but also complex temporal dynamics and inter-frame dependencies.

For AI-generated video attribution, mainstream methods generally cover two categories, including active and passive attribution. Active attribution methods~\cite{Videoshield,Videomark,water-post-1,water-post-2,water-post-3,water-post-4} embed the model's ownership information into the video, which is a watermark-based method. Although this is effective for video attribution, it requires complex operations during watermark embedding, which may affect the visual quality of the video, and not all models will responsibly embed identification watermarks, thus showing significant limitations. The vast majority of passive attribution methods are based on training. VGMShield~\cite{VGM} utilizes a masked autoencoder as the backbone network to capture spatial-temporal dynamic inconsistencies between video frames, significantly improving attribution performance. Vahdati~\textit{et al.}~\cite{vahdati2024beyond} deeply reveal the fundamental reason why AI-generated image detectors are not suitable for detecting AI-generated videos: there is an essential difference in the forensic traces exposed by images and videos, and complete the video attribution in close-set scenarios by learning the unique traces of videos. SAGA~\cite{kundu2025sagasourceattributiongenerative} pioneers a multi-granularity attribution system that includes five dimensions, which solves the problem of precise video attribution where annotated data is extremely scarce. Though several works~\cite{wang2026swiftslidingwindowreconstruction} begin to focus on training-free video attribution tasks, they still have limitations in terms of applicable models and robustness to interferences. In contrast, we propose a training-free video attribution method with strong capabilities of generalization and robustness.

\section{Method}
We view AI-generated video attribution as a retrieval task rather than a traditional classification task, which naturally supports open-set scenarios. This paradigm offers two key advantages. First, it is highly scalable, as newly emerging generators can be incorporated by simply registering new samples without retraining the model. Second, it enables evidence-based attribution, where retrieved reference videos provide intuitive support for the predicted results and thus enhance their interpretability and reliability.
An overview of our approach is shown in Fig.~\ref{fig:method}.
Specifically, we maintain a reference database of AI-generated videos, denoted as $\mathcal{D} = \{v^j_1, \ldots, v^j_i, \ldots, v^J_N\}$, where $j$ denotes the index of generator $G_j$ and $i$ denotes the index of a video sample. We regard generator-specific fingerprints embedded in video frames as the key basis for video attribution. Accordingly, each video in the database is processed to extract fingerprint-aware features, as detailed in Secs.~\ref{sec:3.1}--\ref{sec:3.3}, which are then projected into a unified feature space by a pre-trained video encoder. For a query video, we apply the same processing pipeline and perform attribution via retrieval in the reference database.


\begin{figure*}[t]
    \centering
    \includegraphics[width=\linewidth]{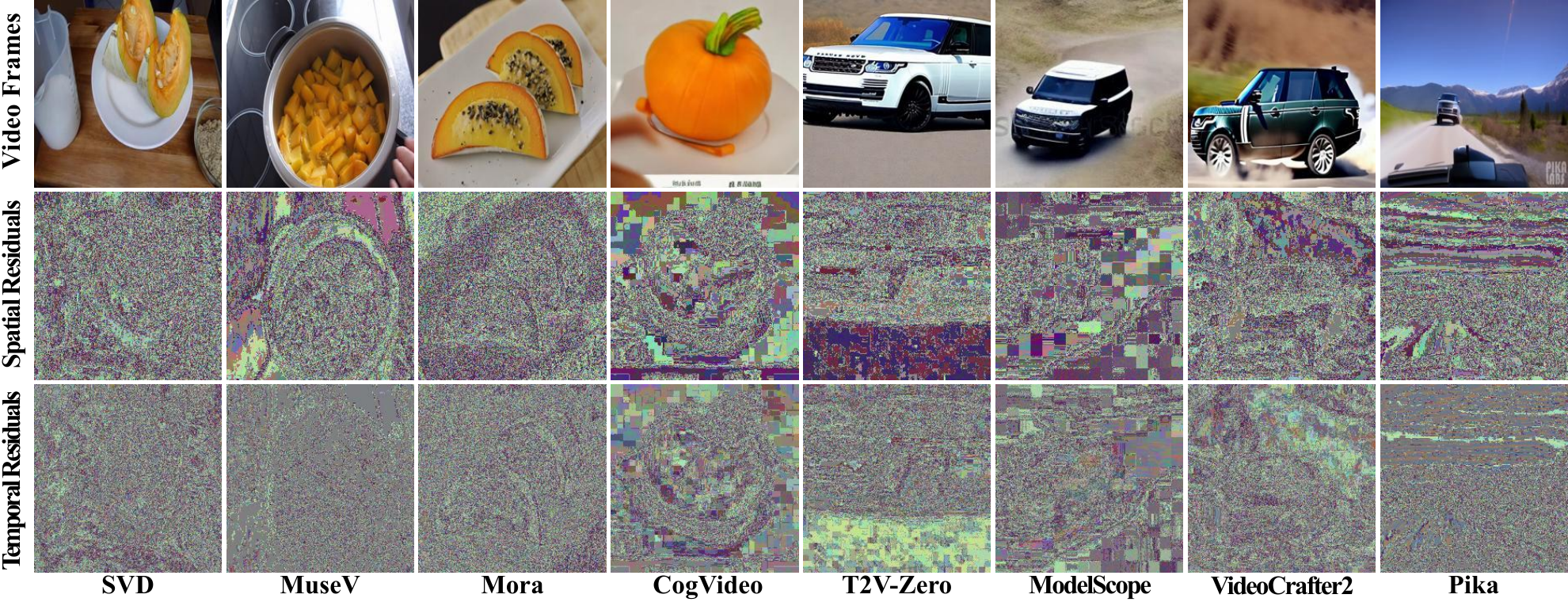}
    \caption{Visualizations of residual frames from different generative models. We visualize the raw video frames, spatial residual frames, and temporal residual frames from eight generative models. From the first column to the eighth column, the first four groups and the last four groups of video frames are each generated from the same semantic information.}
    \label{fig:residual}
\end{figure*}

\begin{figure}[t]
    \centering
    \includegraphics[width=\linewidth]{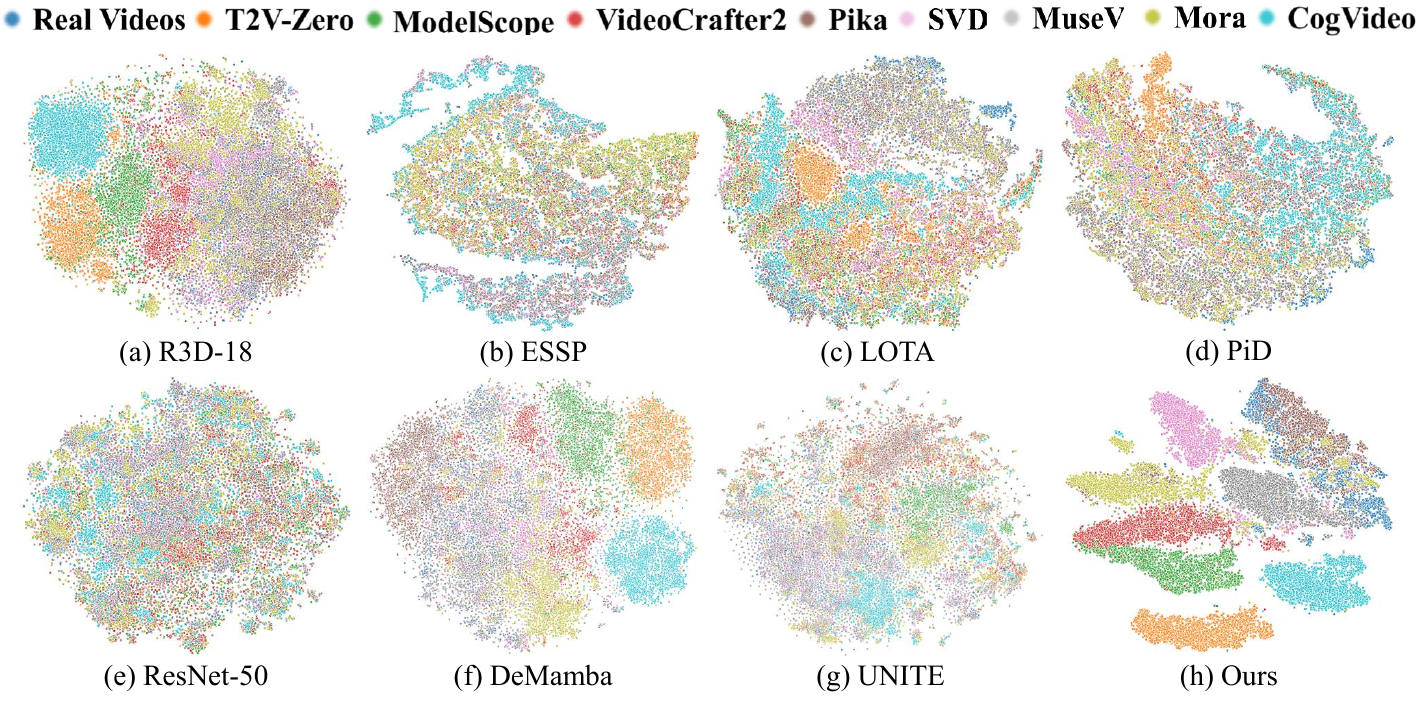}
    \caption{Visualization of video features distribution from real and eight generative sources by using our approach and other mainstream methods.}
    \label{fig:feature}
\end{figure}

\subsection{Adapted Orthogonal Color Transformation}
\label{sec:3.1}
Image color space transformation applies a mapping from the RGB space to a new coordinate system, where different axes correspond to interpretable or task-relevant color components. Unlike fixed color spaces that rely on predefined coefficients, we introduce an adaptive color transformation that adjusts to the video frame.

However, an RGB video frame is dominated by low-frequency components and fails to capture informative local structures. To address this limitation, we propose computing the transformation based on high-frequency residuals, obtained by subtracting a local average from the frame. 
Furthermore, we impose simplex and channel-wise constraints on the transformation 
to ensure perceptual consistency and stability, leading to the Adapted Orthogonal Color Transformation (AOCT) that balances adaptivity and robustness. The AOCT is formulated as:
\begin{align}
\max_{M \in \mathbb{R}^{3 \times 3}} \quad 
& f(\mathbf{I}, M) = \left\| \big(\mathbf{I} - \mathbf{I} \ast K\big) \times_3 M^\top \right\|_F^2 \label{eq:objective} \\
\text{s.t.} \quad 
& M M^\top = I_{3 \times 3}, \label{eq:orthogonality} \\
& \mathbf{m}_1 \in \{ \mathbf{w} \in \mathbb{R}^3 \mid w_i \ge 0,\ \sum_{i=1}^{3} w_i = 1 \}, \label{eq:simplex} \\
& m_{1,G} \ge \tau. \label{eq:green_constraint}
\end{align}
where $I \in \mathbb{R}^{H \times W \times 3}$ denotes the input RGB video frame, \( K = \frac{1}{9} \mathbf{1}_{3 \times 3} \) denotes a $3 \times 3$ mean filter applied channel-wise, $*$ denotes channel-wise convolution, $\times_3$ represents the mode-3 tensor-matrix product along the channel dimension, $M \in \mathbb{R}^{3 \times 3}$ is the learnable transformation matrix, and $\|\cdot\|_F$ denotes the Frobenius norm of a tensor.

Although the transformation matrix $M$ contains nine parameters, it has only two independent degrees of freedom under the imposed constraints. The objective function in Eq.~(\ref{eq:objective}) admits a closed-form solution, and the corresponding independent parameter $\mathbf{m}_1$ can be computed as follows:
\begin{align}
\mathbf{E} &= \mathbf{I} - \mathbf{I} \ast K, \label{eq:residual_matrix} \\
\Sigma &= \frac{1}{N} E^\flat (E^\flat)^\top, \label{eq:cov_matrix} \\
\mathbf{m}_1^\star &= \arg\max_{\|\mathbf{m}\|_2 = 1} 
\mathbf{m}^\top \Sigma \mathbf{m} = \mathbf{v}_{\max}(\Sigma),\label{eq:eigen_matrix}
\end{align}
where $E$ is the intermediate residual image, $E^\flat$ denotes the flattened residual matrix, obtained by reshaping the $E$ such that each column corresponds to the RGB vector of a single pixel.

After obtaining the independent parameter $\mathbf{m}_1$, the transformation matrix $M$ is computed as:
\begin{align}
\tilde{\mathbf{m}}_1 &= \Pi_{\mathcal{C}} \left( \mathbf{m}_1^\star \right), \label{eq:projection_matrix} \\
M^\star &= \mathrm{GS}\left( \tilde{\mathbf{m}}_1 \right). \label{eq:gs_matrix}
\end{align}
where $\tilde{\mathbf{m}}_1$ denotes the projected independent vector under imposed constraints, and $\mathrm{GS}$ represents the Gram-Schmidt orthogonalization that constructs an orthonormal basis. The proof of related theories and propositions can be found in the \textit{Supplementary Material}.

Given the input frame $\mathbf{I}$, the AOCT-transformed frame $\mathbf{Y}$ is:
\begin{equation}
    \mathbf{Y} = \mathbf{I} \times_3 M.
\end{equation}

\subsection{Multi-Scale Quantized Residual}
\label{sec:3.2}
An RGB video frame is typically dominated by low-frequency components, which can obscure informative local structures. To effectively capture informative local structures in RGB images while suppressing redundant low-frequency components, we propose the Multi-Scale Quantization Residual (MSQR) of the color-transformed frame. At quantization scale $i$, we quantize each channel as:
\begin{equation}
    \hat{\mathbf{Y}}^{(i)} = \operatorname{round}\Big( \mathbf{Y} \oslash \mathbf{S}^{(i)} \Big) \odot \mathbf{S}^{(i)},
\end{equation}
where $\mathbf{S}^{(i)} \in \mathbb{R}^{1 \times 1 \times 3}$ is the scale vector for each channel, $\oslash$ denotes element-wise division, and $\odot$ denotes element-wise multiplication. This operation discretizes the transformed image at multiple scales, thereby preserving both coarse and fine structures.

To extract informative differences between the original and quantized frames, we compute the residual at each scale by inverting the color transformation:
\begin{equation}
    \mathbf{R}^{(i)} = \mathbf{I} - M^{-1} \hat{\mathbf{Y}}^{(i)}.
\end{equation}

Finally, the multi-scale residual is obtained by averaging over all $n$ quantization scales:
\begin{equation}
    \mathbf{R} = \frac{1}{n} \sum_{i=1}^{n} \mathbf{R}^{(i)}.
\end{equation}

By performing multi-scale quantization, we isolate high-frequency details across different resolutions, enabling more robust residual extraction. The resulting multi-scale frame residual highlights informative variations while suppressing redundant low-frequency content, thereby improving the stability and effectiveness of subsequent color and structural transformations.

\subsection{Temporal-Semantic Aggregation}
\label{sec:3.3}
\label{sec:triple}



For the residual sequence $\mathbf{R} = \{r_1, r_2, \dots, r_T\}$ generated by the MSQG module, where $T$ is the total number of frames, we compute adjacent frame differences $\Delta \mathbf{R} = \{\Delta r_1, \Delta r_2, \dots, \Delta r_T\}$ to capture the instantaneous temporal changes in the residuals across frames:
\begin{equation}
\Delta \mathbf{R} =
\begin{cases}
r_{t+1} - r_t, & 1 \le t < T, \\
r_T - r_{T-1}, & t = T.
\end{cases}
\end{equation}

We finally utilize spatial residual stream ($R$) to capture generative artifacts in spatial domain, employ temporal residual stream ($\Delta R$) to focus on unnatural fluctuations in the temporal dimension, and introduce RGB context stream ($I$) to take global semantic backgrounds into account. The final discriminative feature $F$ is obtained by weighted fusion of three streams:
\begin{equation}
    \mathbf{F} = w_s \cdot \mathbf{R} + w_t \cdot \Delta \mathbf{R} + w_{rgb} \cdot \mathbf{I}
\end{equation}
where $w_s$, $w_t$ and $w_{rgb}$ are different weights of spatial residual stream, temporal residual stream, and RGB context stream. Additionally, $\Delta R$ is the first-order temporal difference of the sequence $R$, with end-frame repeat-padding applied.

Fig.~\ref{fig:residual} presents the raw video frames, spatial residual frames, and temporal residual frames sampled from videos of eight generative models. The former four sets (1)$\sim$(4) shares the same semantic information "\textit{A white plate topped with an orange squash}", and the latter four sets (5)$\sim$(8) are generated by the same semantic prompts "\textit{A car speeding on a wilderness road}".
Despite their similar visual content, videos generated by the eight models exhibit distinct model-specific fingerprints in both spatial and temporal residuals. For example, MuseV tends to leave fingerprint patterns along contour boundaries, whereas Mora more commonly reveals such patterns in intricate texture regions. Pika often induces unnatural fluctuations in background color gradients, while ModelScope is characterized by conspicuous pixelation blocks. These distinctive residual patterns are strongly model-dependent and therefore highly informative for video attribution.
By extracting features from our proposed residuals using a feature extractor, we observe that most categories are clearly separated (Fig.~\ref{fig:feature} (h)), whereas features from the original RGB frames remain mixed (Fig.~\ref{fig:feature} (e)).

\begin{table*}[t]
\begin{adjustbox}{max width=\textwidth}
\begin{tabular}{c|l|cc|cc|cc|cc|cc|cc|cc|cc|cc|cc}
\toprule
\multirow{2}{*}{Shot} & \multirow{2}{*}{Method} & \multicolumn{2}{c|}{Real} & \multicolumn{2}{c|}{T2VZ} & \multicolumn{2}{c|}{MS} & \multicolumn{2}{c|}{VC2} & \multicolumn{2}{c|}{Pika} & \multicolumn{2}{c|}{SVD} & \multicolumn{2}{c|}{MuseV} & \multicolumn{2}{c|}{Mora} & \multicolumn{2}{c|}{CogV} & \multicolumn{2}{c}{Avg.} \\
&  & Rank-1 & mAP & Rank-1 & mAP & Rank-1 & mAP & Rank-1 & mAP & Rank-1 & mAP & Rank-1 & mAP & Rank-1 & mAP & Rank-1 & mAP & Rank-1 & mAP & Rank-1 & mAP \\ \midrule  
\multirow{8}{*}{\rotatebox{90}{1-shot}}    
&ResNet-50~\cite{he2016deep} &10.8 &15.2 &8.0 &9.1 &23.6 &26.5 &29.2 &29.1 &\textbf{39.7} &34.5 &\underline{36.7} &30.1 &6.0 &8.5 &3.5 &6.8 &6.0 &8.7 &17.4 &18.4 \\
&R3D-18~\cite{tran2018closer} &\textbf{65.8} &\textbf{78.8} &29.6 &\underline{50.4} &12.6 &31.7 &28.6 &50.3 &23.1 &\textbf{46.5} &\underline{36.7} &\underline{58.8} &1.5 &\underline{24.1} &7.0 &24.2 &\underline{52.3} &71.1 &\underline{28.6} &\underline{48.4} \\
&ESSP~\cite{chen2024single} &9.6 &5.0 &3.5 &5.3 &\underline{39.2} &\underline{36.8} &\underline{42.7} &42.1 &7.0 &9.4 &8.5 &8.2 &4.0 &2.8 &\underline{42.7} &\underline{46.1} &8.5 &21.7 &18.4 &19.7 \\
&LOTA~\cite{Wang_2025_ICCV} &\underline{30.2} &\underline{43.9} &13.6 &24.6 &\textbf{96.5} &\textbf{87.9} &\textbf{59.3} &\textbf{63.9} &11.1 &20.6 &1.6 &7.5 &4.0 &8.4 &21.6 &36.5 &27.6 &39.6 &29.5 &37.0 \\
&PiD~\cite{Fu_2025_ICML} &9.1 &10.9 &13.6 &14.5 &6.0 &9.2 &26.6 &45.9 &\underline{36.7} &\underline{45.5} &22.6 &38.8 &8.5 &19.6 &31.7 &42.6 &\textbf{62.8} &\textbf{69.7} &24.2 &32.9 \\
&DeMamba~\cite{DeMamba} &2.0 &- &\textbf{66.3} &- &9.6 &- &16.8 &- &33.7 &- &22.1 &- &5.5 &- &32.2 &- &29.2 &- &24.1 &- \\
&UNITE~\cite{kundu2025universalsyntheticvideodetector} &2.8 &- &0.5 &- &0.5 &- &1.0 &- &0.2 &- &52.3 &- &\textbf{62.}8 &- &3.0 &- &1.5 &- &13.8 &- \\
&Ours &8.5 &28.2 &\underline{56.3} &\textbf{66.1} &20.6 &36.1 &28.6 &\underline{56.0} &11.6 &42.1 &\textbf{40.7} &\textbf{61.9} &\underline{52.3} &\textbf{69.4} &\textbf{54.3} &\textbf{66.0} &22.1 &\underline{42.3} &\textbf{32.8} &\textbf{52.0} \\ \midrule
\multirow{8}{*}{\rotatebox{90}{10-shot}} 
&ResNet-50~\cite{he2016deep} &33.4 &28.9 &13.6 &14.8 &26.1 &20.7 &15.6 &18.6 &29.2 &23.0 &15.6 &17.3 &16.1 &18.7 &17.1 &17.6 &41.7 &31.5 &24.2 &32.0 \\
&R3D-18~\cite{tran2018closer} &17.1 &25.5 &45.2 &45.8 &28.1 &34.6 &30.2 &37.3 &43.7 &44.1 &21.6 &31.5 &65.3 &\underline{63.1} &12.6 &22.5 &\underline{84.4} &\underline{76.6} &38.7 &42.3 \\
&ESSP~\cite{chen2024single} &18.6 &16.2 &16.1 &16.8 &59.8 &50.0 &31.2 &25.3 &27.6 &20.5 &10.1 &12.8 &19.1 &19.9 &12.6 &18.0 &37.2 &30.4 &25.8 &23.3 \\
&LOTA~\cite{Wang_2025_ICCV} &34.2 &\underline{38.0} &52.8 &\underline{48.1} &\textbf{78.4} &\textbf{74.9} &\underline{51.3} &\underline{51.5} &\underline{41.2} &42.2 &\underline{38.2} &35.9 &40.2 &44.4 &29.6 &32.0 &81.4 &63.5 &\underline{49.7} &\underline{47.8} \\
&PiD~\cite{Fu_2025_ICML} &34.2 &37.4 &43.7 &44.5 &38.7 &39.1 &26.6 &28.6 &\textbf{50.3} &\underline{44.8} &36.2 &\underline{39.4} &40.7 &41.8 &\underline{38.7} &\underline{38.3} &67.8 &52.6 &41.9 &40.7 \\
&DeMamba~\cite{DeMamba} &12.6 &- &\underline{60.8} &- &22.6 &- &24.6 &- &27.1 &- &22.6 &- &14.1 &- &11.6 &- &29.6 &- &25.1 &- \\
\textbf{}&UNITE~\cite{kundu2025universalsyntheticvideodetector} &\underline{47.0} &- &48.2 &- &26.1 &- &27.1 &- &18.6 &- &14.1 &- &\underline{72.9} &- &37.2 &- &76.5 &- &40.8 &- \\
&Ours &\textbf{61.8} &\textbf{57.3} &\textbf{99.0} &\textbf{94.8} &\underline{50.7} &\underline{54.7} &\textbf{54.7} &\textbf{57.8} &41.2 &\textbf{50.3} &\textbf{76.4} &\textbf{73.4} &\textbf{84.9} &\textbf{76.8} &\textbf{61.8} &\textbf{58.8} &\underline{80.9} &\textbf{81.6} &\textbf{68.0} &\textbf{67.3} \\ \midrule
\multirow{8}{*}{\rotatebox{90}{100-\textbf{}shot}}    
&ResNet-50~\cite{he2016deep} &34.4 &32.9 &22.6 &21.5 &39.2 &31.4 &27.6 &24.2 &27.6 &24.7 &23.1 &19.6 &12.6 &14.8 &22.1 &20.6 &45.7 &34.8 &28.9 &25.8 \\
&R3D-18~\cite{tran2018closer} &43.0 &45.4 &73.4 &66.1 &47.7 &48.9 &42.7 &42.9 &53.8 &53.4 &21.1 &27.2 &48.7 &49.6 &36.2 &38.4 &\underline{92.0}
&\underline{87.8} &50.9 &51.1 \\
&ESSP~\cite{chen2024single} &21.6 &20.2 &14.6 &15.6 &55.3 &49.4 &28.6 &26.1 &14.1 &18.1 &21.6 &20.4 &18.1 &18.2 &23.1 &22.6 &38.7 &34.6 &26.2 &25.0 \\
&LOTA~\cite{Wang_2025_ICCV} &\underline{59.3} &\underline{54.9} &\underline{79.0} &\underline{70.3} &\textbf{96.0} &\textbf{92.9} &\underline{64.3} &\underline{58.9} &\underline{55.8} &\underline{57.6} &46.2 &46.5 &50.3 &51.3 &40.7 &41.9 &85.9 &81.2 &\underline{64.1} &\underline{61.7} \\
&PiD~\cite{Fu_2025_ICML} &44.2 &45.2 &70.4 &63.5 &58.3 &54.7 &39.7 &42.8 &\underline{55.8} &54.6 &\underline{54.3} &\underline{51.3} &51.3 &52.8 &\underline{42.2} &\underline{43.9} &88.9 &81.1 &56.1 &54.4 \\
&DeMamba~\cite{DeMamba} &17.6 &- &61.3 &- &35.7 &- &36.7 &- &40.7 &- &21.1 &- &28.6 &- &23.1 &- &37.7 &- &33.6 &- \\
&UNITE~\cite{kundu2025universalsyntheticvideodetector} &5.3 &- &88.9 &- &79.4 &- &15.6 &- &66.3 &- &8.0 &- &56.3 &- &85.4 &- &85.4 &- &49.6 &- \\
&Ours &\textbf{60.5} &\textbf{58.1} &\textbf{97.5} &\textbf{96.7} &\underline{82.9} &\underline{76.2} &\textbf{91.5} &\textbf{77.5} &\textbf{73.4} &\textbf{65.7} &\textbf{82.9} &\textbf{78.1} &\textbf{91.9} &\textbf{86.0} &\textbf{80.9} &\textbf{71.4} &\textbf{99.5} &\textbf{95.0} &\textbf{84.6} &\textbf{78.3} \\ \bottomrule
\end{tabular}
\end{adjustbox}
\caption{Comparison of attribution performance of our approach against competing methods under 1-shot, 10-shot and 100-shot, respectively. The best results are highlighted in bold, and the second-best results are underlined.}

\label{tab:comparison}
\end{table*}

\begin{table}[t]
\begin{adjustbox}{max width=\linewidth}
\begin{tabular}{c|l|ccccccccccc}
\toprule
Shot &Method &HD-VG &Vript &T2VZ &MS &VC2 &Pika &SVD &MuseV &Mora &CogV & Avg. \\ \midrule
\multirow{8}{*}{\rotatebox{90}{1-\textbf{}shot}}    
&ResNet-50~\cite{he2016deep} &33.1 &\underline{52.4} &52.3 &59.3 &43.2 &43.2 &59.8 &63.8 &41.7 &50.8 &\cellcolor{gray!35}50.8 \\
&R3D-18~\cite{tran2018closer} &\textbf{70.4} &\textbf{67.3} &97.5 &91.4 &74.9 &44.7 &52.7 &27.6 &32.2 &99.5 &\cellcolor{gray!35}65.8 \\
&ESSP~\cite{chen2024single} &11.1 &10.6 &84.4 &85.4 &79.9 &96.0 &93.4 &94.0 &92.4 &89.5 &\cellcolor{gray!35}71.7 \\
&LOTA~\cite{Wang_2025_ICCV} &10.5 &16.6 &\underline{99.5} &\textbf{100.0} &\textbf{100.0} &89.4 &95.6 &98.5 &95.0 &99.3 &\cellcolor{gray!35}80.4 \\
&PiD~\cite{Fu_2025_ICML} &20.1 &10.6 &\textbf{100.0} &\underline{99.6} &\textbf{100.0} &\underline{96.8} &\textbf{99.4} &98.1 &96.0 &\underline{99.0} &\cellcolor{gray!35}\underline{82.0} \\
&DeMamba~\cite{DeMamba} &\underline{67.8} &46.7 &63.3 &65.8 &61.3 &56.3 &32.2 &36.7 &48.2 &31.2 &\cellcolor{gray!35}50.9 \\
&UNITE~\cite{kundu2025universalsyntheticvideodetector} &0 &0 &\textbf{100.0} &99.3 &\underline{99.2} &\textbf{100.0} &\underline{99.0} &\textbf{99.6} &\textbf{100.0} &99.9 &\cellcolor{gray!35}79.7 \\ 
&Ours &15.1 &12.6 &\textbf{100.0} &\textbf{100.0} &\textbf{100.0} &\textbf{100.0} &\underline{99.0} &\underline{99.5} &\textbf{100.0} &98.8 &\cellcolor{gray!35}\textbf{82.5} \\ \midrule
\multirow{8}{*}{\rotatebox{90}{10-\textbf{}shot}}    
&ResNet-50~\cite{he2016deep}                           &34.0 &41.4 &86.4 &83.4 &88.4 &74.9 &65.8 &69.4 &83.4 &89.5 &\cellcolor{gray!35}75.4 \\
&R3D-18~\cite{tran2018closer}                          &23.6 &25.1 &96.5 &86.9 &83.4 &85.4 &71.9 &83.4 &89.5 &98.5 &\cellcolor{gray!35}74.4 \\
&ESSP~\cite{chen2024single}                            &20.6 &17.1 &93.5 &92.9 &86.9 &83.8 &95.1 &87.4 &94.5 &88.4 &\cellcolor{gray!35}76.0 \\
&LOTA~\cite{Wang_2025_ICCV}                            &36.2 &41.2 &\textbf{100.0} &99.0 &\underline{99.5} &88.4 &89.4 &84.9 &96.8 &\textbf{100.0} &\cellcolor{gray!35}83.5 \\
&PiD~\cite{Fu_2025_ICML}                               &33.7 &\underline{42.7} &\underline{99.6} &\underline{99.4} &\underline{99.5} &\underline{90.9} &96.1 &\underline{97.2} &\textbf{98.8} &98.5 &\cellcolor{gray!35}\underline{85.7} \\
&DeMamba~\cite{DeMamba}                                &\underline{36.7} &35.2 &99.1 &98.8 &91.9 &88.0 &64.8 &58.3 &86.4 &94.0 &\cellcolor{gray!35}75.3 \\
&UNITE~\cite{kundu2025universalsyntheticvideodetector} &2.0 &0.5 &\textbf{100.0} &\textbf{100.0} &\textbf{100.0} &\textbf{100.0} &\textbf{99.4} &\textbf{99.1} &\underline{98.6} &\underline{99.4} &\cellcolor{gray!35}79.9 \\ 
&Ours                                                       &\textbf{49.3} &\textbf{54.8} &\textbf{100.0} &\textbf{100.0} &\textbf{100.0} &79.0 &\underline{97.5} &95.6 &90.4 &\textbf{100.0} &\cellcolor{gray!35}\textbf{86.6} \\ \midrule
\multirow{8}{*}{\rotatebox{90}{100-\textbf{}shot}}    
&ResNet-50~\cite{he2016deep}                           &19.0 &35.4 &88.9 &84.8 &87.7 &86.6 &83.0 &82.4 &87.9 &90.9 &\cellcolor{gray!35}80.0 \\
&R3D-18~\cite{tran2018closer}                          &32.2 &42.2 &96.5 &92.0 &88.9 &86.4 &68.3 &79.4 &77.8 &\textbf{100.0} &\cellcolor{gray!35}76.4 \\
&ESSP~\cite{chen2024single}                            &26.1 &21.6 &85.9 &92.3 &86.1 &83.4 &85.9 &82.4 &90.4 &81.4 &\cellcolor{gray!35}73.6 \\
&LOTA~\cite{Wang_2025_ICCV}                            &\textbf{57.8} &53.8 &\underline{99.5} &\textbf{100.0} &99.4 &87.4 &94.0 &91.9 &89.1 &\textbf{100.0} &\cellcolor{gray!35}87.3 \\
&PiD~\cite{Fu_2025_ICML}                               &50.7 &\underline{65.3} &\textbf{100.0} &\textbf{100.0} &\textbf{100.0} &93.0 &\underline{99.1} &\underline{98.4} &95.3 &\underline{99.8} &\cellcolor{gray!35}\underline{90.2} \\
&DeMamba~\cite{DeMamba}                                &54.8 &59.3 &94.0 &\underline{96.1} &\underline{95.4} &\underline{96.5} &62.7 &41.2 &82.9 &90.9 &\cellcolor{gray!35}76.4 \\
&UNITE~\cite{kundu2025universalsyntheticvideodetector} &13.6 &12.6 &\textbf{100.0} &\textbf{100.0} &\textbf{100.0} &\textbf{100.0} &98.5 &\textbf{99.0} &\textbf{99.4} &98.6 &\cellcolor{gray!35}82.2 \\ 
&Ours                                                       &\underline{55.3} &\textbf{67.8} &\textbf{100.0} &\textbf{100.0} &\textbf{100.0} &93.0 &\textbf{99.5} &96.1 &\underline{98.9} &99.5 &\cellcolor{gray!35}\textbf{91.0} \\ \bottomrule
\end{tabular}
\end{adjustbox}
\caption{Comparison of detection accuracy of our approach against competing methods under 1-shot, 10-shot and 100-shot, respectively. 
}
\label{tab:deepfake}
\end{table}

\begin{table*}[t]
    \centering
    \begin{minipage}[t]{\columnwidth} 
        \centering
        \adjustbox{max width=\textwidth}{
        \begin{tabular}{c|cccc|cccccccccc}
        \toprule
        Row & AOCT &MSQR &Temporal &RGB &Real &T2VZ &MS &VC2 &Pika &SVD &MuseV &Mora &CogV & Avg. \\ \midrule
        1 & \ding{55} & \ding{55} & \ding{55} &\ding{55} &45.4 &66.1 &48.9 &42.9 &53.4 &27.2 &49.6 &38.4 &87.8 &\cellcolor{gray!35}51.1 \\
        2 & \ding{51} & \ding{55} & \ding{55} &\ding{55} &49.6 &77.3 &68.1 &59.1 &55.7 &54.1 &65.3 &47.8 &80.6 &\cellcolor{gray!35}62.0 \\
        3 & \ding{51} & \ding{51} & \ding{55} &\ding{55} &51.4 &86.9 &72.1 &56.3 &60.9 &59.6 &73.7 &49.7 &90.9 &\cellcolor{gray!35}66.8 \\
        4 & \ding{51} & \ding{55} & \ding{51} &\ding{55} &53.5 &93.7 &75.0 &66.3 &\textbf{66.6} &76.1 &77.8 &\underline{66.3} &88.9 &\cellcolor{gray!35}73.8 \\
        5 & \ding{51} & \ding{51} & \ding{51} &\ding{55} &\underline{55.9} &\textbf{97.4} &\textbf{80.2} &\underline{71.8} &60.7 &\textbf{82.1} &\underline{85.7} &65.9 &\underline{92.8} &\cellcolor{gray!35}\underline{76.9} \\
        6 & \ding{51} & \ding{51} & \ding{51} &\ding{51} &\textbf{58.1} &\underline{96.7} &\underline{76.2} &\textbf{77.5} &\underline{65.7} &\underline{78.1} &\textbf{86.0} &\textbf{71.4} &\textbf{95.0} &\cellcolor{gray!35}\textbf{78.3} \\ \bottomrule
        \end{tabular}
        }
        \caption{Ablation studies on effectiveness of each module of our apporach, including AOCT, MSQR, temporal stream (Temporal), and RGB stream (RGB).}
       \label{tab:ablation}
    \end{minipage}
    \hfill
    \begin{minipage}[t]{\columnwidth} 
                \centering
        \adjustbox{max width=\textwidth}{
        \begin{tabular}{c|cccc|cccc}
        \toprule
        Method &H.264-0 &H.264-12 &H.264-18 &H.264-24 &Crop-100 &Crop-90 &Crop-70 &Crop-50 \\ \midrule
        Ours                            &\textbf{84.6} &\textbf{72.7} &\textbf{71.8} &\textbf{71.1} &\textbf{84.6} &\textbf{76.8} &\textbf{73.9} &\textbf{67.8} \\
        ESSP~\cite{chen2024single} &26.2 &19.8 &18.6 &17.4 &26.2 &19.0 &17.8 &17.3 \\
        LOTA~\cite{Wang_2025_ICCV} &\underline{64.1} &\underline{56.8} &\underline{52.4} &\underline{36.8} &\underline{64.1} &\underline{60.7} &\underline{52.3} &\underline{37.7} \\
        DeMamba~\cite{DeMamba}     &33.6 &20.8 &17.7 &12.7 &33.6 &31.8 &32.1 &29.8 \\
        \bottomrule
        \end{tabular}
        }
        \caption{Analysis of robustness to video degradation under H.264 compression (CRF = 12, 18, 24) and central cropping (retaining 90\%, 70\%, and 50\% of the frame).}

        \label{tab:robustness}
    \end{minipage}
    \hfill
    \begin{minipage}[t]{\columnwidth} 
                \centering
        \adjustbox{max width=\textwidth}{
        \begin{tabular}{c|cccccccccc}
        \toprule
        Weight &Real &T2VZ &MS &VC2 &Pika &SVD &MuseV &Mora &CogV & Avg. \\ \midrule
        0.1 &51.3 &\textbf{98.6} &76.1 &\underline{71.2} &62.4 &\textbf{80.2} &\underline{85.5} &61.3 &92.3 &\cellcolor{gray!35}75.4 \\
        0.2 &\textbf{59.7} &\underline{98.5} &\textbf{79.6} &64.3 &\underline{64.8} &\textbf{80.2} &83.2 &64.8 &\underline{94.9} &\cellcolor{gray!35}\underline{76.7} \\
        0.3 &\underline{58.1} &96.7 &76.2 &\textbf{77.5} &\textbf{65.7} &78.1 &\textbf{86.0} &\textbf{71.4} &\textbf{95.0} &\cellcolor{gray!35}\textbf{78.3} \\
        0.4 &57.4 &96.0 &77.0 &68.1 &63.5 &\underline{78.2} &83.3 &\underline{67.1} &92.5 &\cellcolor{gray!35}75.9 \\
        0.5 &56.4 &97.2 &76.1 &66.8 &61.6 &73.5 &81.9 &62.4 &93.8 &\cellcolor{gray!35}74.4 \\
        0.6 &52.0 &93.8 &\underline{77.3} &61.8 &\underline{64.8} &75.4 &83.6 &61.1 &92.6 &\cellcolor{gray!35}73.6 \\ \bottomrule
        \end{tabular}
        }
        \caption{Analysis of the temporal residual stream weight under different settings (0.1–0.6).}
        \label{tab:temporal_weights}
    \end{minipage}
    \hfill
    \begin{minipage}[t]{\columnwidth} 
                \centering
        \adjustbox{max width=\textwidth}{
        \begin{tabular}{c|cccccccccc}
        \toprule
        Weight &Real &T2VZ &MS &VC2 &Pika &SVD &MuseV &Mora &CogV & Avg. \\ \midrule
        0.05 &58.0 &\textbf{97.6} &\textbf{77.8} &70.5 &61.5 &\textbf{81.1} &\textbf{86.9} &63.1 &94.5 &\cellcolor{gray!35}\underline{76.8} \\
        0.10 &58.1 &96.7 &\underline{76.2} &\textbf{77.5} &\textbf{65.7} &\underline{78.1} &\underline{86.0} &\textbf{71.4} &\underline{95.0} &\cellcolor{gray!35}\textbf{78.3} \\
        0.15 &\underline{58.6} &96.6 &72.7 &\underline{73.7} &61.9 &77.5 &84.0 &\underline{67.1} &93.4 &\cellcolor{gray!35}76.2 \\
        0.20 &56.2 &95.3 &73.2 &72.6 &64.2 &73.6 &84.0 &65.9 &93.5 &\cellcolor{gray!35}75.4 \\
        0.25 &\textbf{59.0} &\underline{97.0} &71.5 &73.5 &\underline{65.4} &71.3 &83.9 &63.5 &\textbf{96.2} &\cellcolor{gray!35}75.7 \\
        0.30 &52.6 &95.1 &67.1 &65.5 &66.3 &66.9 &79.8 &63.0 &93.8 &\cellcolor{gray!35}72.2 \\ \bottomrule
        \end{tabular}
        }
        \caption{Analysis of the RGB context stream weight under different settings (0.05--0.30).}
        \label{tab:rgb_weights}
    \end{minipage}
    \hfill
    \begin{minipage}[t]{\columnwidth} 
        \centering
        \adjustbox{max width=\textwidth}{
        \begin{tabular}{cc|cccccccccc}
        \toprule
        Order &Matrix &Real &T2VZ &MS &VC2 &Pika &SVD &MuseV &Mora &CogV & Avg. \\ \midrule
        (1) &UDM        &55.7 &95.4 &\underline{77.8} &\underline{80.1} &63.7 &75.4 &77.2 &62.8 &89.2 &\cellcolor{gray!35}75.2 \\
        (2) &DCT        &53.6 &96.1 &68.7 &67.7 &\underline{74.2} &77.2 &84.3 &67.6 &89.8 &\cellcolor{gray!35}75.5 \\ 
        (3) &YCbCr      &\underline{56.5} &96.0 &71.8 &72.3 &\textbf{76.0} &73.2 &83.4 &65.5 &93.2 &\cellcolor{gray!35}76.4 \\
        (4) &YUV        &53.9 &95.2 &72.8 &74.6 &73.8 &79.4 &84.7 &64.1 &\underline{94.0} &\cellcolor{gray!35}76.9 \\
        (5) &R-Adaptive &52.3 &\underline{96.5} &76.2 &\textbf{80.2} &59.7 &\textbf{81.8} &84.0 &69.3 &93.1 &\cellcolor{gray!35}77.0 \\
        (6) &B-Adaptive &54.6 &96.0 &\textbf{82.1} &\underline{80.1} &58.9 &\underline{79.7} &\underline{85.0} &\underline{71.2} &\underline{94.0} &\cellcolor{gray!35}\underline{78.0} \\
        (7) &Ours       &\textbf{58.1} &\textbf{96.7} &76.2 &77.5 &65.7 &78.1 &\textbf{86.0} &\textbf{71.4} &\textbf{95.0} &\cellcolor{gray!35}\textbf{78.3} \\
        \bottomrule
        \end{tabular}
        }
        \caption{Analysis of the transformation matrix. We compare our adaptive matrix with predefined and alternative adaptive matrices, including UDM, DCT, YCbCr, YUV, R-Adaptive, and B-Adaptive.}

        \label{tab:matrix}
    \end{minipage}
    \hfill
    \begin{minipage}[t]{\columnwidth} 
                \centering
        \adjustbox{max width=\textwidth}{
        \begin{tabular}{c|cc|c}
        \toprule
        Method &Preparation Time &Attribution Time &Total Time \\ \midrule
        ESSP~\cite{chen2024single} &1392.25 s &1422.76 s &2815.01 s \\
        LOTA~\cite{Wang_2025_ICCV} &\underline{999.71 s} &\underline{1018.14 s} &\underline{2017.85 s} \\
        DeMamba~\cite{DeMamba} &2167.39 s &1706.85 s &3874.24 s \\
        UNITE~\cite{kundu2025universalsyntheticvideodetector} &1483.46 s &1681.33 s &3164.79 s \\
        Ours  &\textbf{659.73 s} &\textbf{672.57 s} &\textbf{1332.30 s} \\
        \bottomrule
        \end{tabular}
        }
        \caption{Analysis of the efficiency of video attribution. We compare the time consumption of our approach with mainstream methods, including both preparation time (database construction or training) and attribution time.}

        \label{tab:efficiency}
    \end{minipage}
\end{table*}

\section{Experiments}

\subsection{Experimental Setup}

\noindent\textbf{Dataset:} GenVidBench~\cite{ni2025genvidbenchchallengingbenchmarkdetecting} is a comprehensive and recently introduced benchmark for AI-generated video detection. It contains 100,000 semantic labels, along with the original prompts and images used during the generation process. The dataset combines real-world videos from HD-VG~\cite{hd_vg_130m} and Vript~\cite{vript} with synthetic videos produced by eight generators: T2V-Zero (T2VZ)~\cite{t2vz}, ModelScope (MS)~\cite{modelscope}, VideoCrafter2 (VC2)~\cite{Videocrafter2}, Pika~\cite{pika}, SVD~\cite{blattmann2023stable}, MuseV~\cite{musev}, Mora~\cite{mora}, and CogVideo (CogV)~\cite{cogvideo}. Since it covers both AI-generated videos from diverse mainstream generators and semantically related real videos, it is well suited for AI-generated video attribution. In our setting, we combine HD-VG~\cite{hd_vg_130m} and Vript~\cite{vript} into a single \textit{Real} category.

\noindent\textbf{Evaluation Metrics:} For AI-generated video attribution, we report Rank-1 and mean Average Precision (mAP). Rank-1 measures the proportion of query videos whose top prediction matches the ground-truth source, while mAP computes the mean average precision over all query videos. For AI-generated video detection, we use accuracy as the evaluation metric.

\noindent\textbf{Implementation Details:} We sample 64 consecutive frames from each video and resize them to 224×224. Videos with fewer than 64 frames are padded by temporal looping. In the adaptive color space, the quantization factors for the first channel are set to [0.50, 1.20, 1.90, 2.60], while those for remaining channels are set to [0.40, 1.00, 1.60, 2.20]. The resulting frames obtained under different quantization factors are then averaged. During feature fusion, the weights of the spatial residual, temporal residual, and RGB context streams are set to 0.6, 0.3, and 0.1, respectively. The fused features are then fed into an R3D-18 backbone pre-trained on Kinetics-400 for feature extraction. The extracted features are subsequently stored in a reference database. During retrieval, query videos undergo the same process, and cosine similarity is computed against all stored features for source attribution or deepfake detection.

\subsection{Experimental Results} 
To comprehensively evaluate our approach, we compare it against baselines of AI-generated image detection and attribution: (1) \textbf{Constructed baselines}: We employ ResNet-50~\cite{he2016deep}, a CNN-based model designed for images, to process each frame of a video, and the features generated from all frames are averaged to construct the database. We also utilize R3D-18~\cite{tran2018closer}, a CNN-based model designed for videos, to process a series of frames to generate features for the database. (2) \textbf{AI-generated image detection methods}: The input frames are first processed by the image processing method proposed in ESSP~\cite{chen2024single}, LOTA~\cite{Wang_2025_ICCV} or PiD~\cite{Fu_2025_ICML}, and then are input into R3D-18 to obtain features for constructing the database. (3) \textbf{AI-generated video attribution methods}: We train DeMamba~\cite{DeMamba} and UNITE~\cite{kundu2025universalsyntheticvideodetector} with the classifier replaced by a nine-class classifier on randomly selected videos from subsets of GenVidBench, and evaluate them on all videos of GenVidBench.\\

\noindent\textbf{Attribution Performance:} We evaluate the attribution performance of our approach under three settings: 1-shot, 10-shot, and 100-shot, corresponding to registering 1, 10, and 100 video features per source in the database, respectively. 
Results in Tab.~\ref{tab:comparison} show that our approach outperforms all competing methods. In particular, in the 100-shot setting, it achieves 84.6\% Rank-1 and 78.3\% mAP, surpassing competitors by over 20.5\% Rank-1 and 16.6\% mAP.
Even in the case of 1-shot, our approach maintains a reasonable mAP of 52.0$\%$. It should be noted that other methods struggle with a huge decline on certain classes, while our approach achieves consistently excellent results in various classes. As shown in Fig.~\ref{fig:feature}, deepfake detetcion methods (ResNet-50~\cite{he2016deep}, ESSP~\cite{chen2024single}, LOTA~\cite{Wang_2025_ICCV}, PiD~\cite{Fu_2025_ICML}) are difficult to generalize to video attribution tasks, while deepfake video detection methods (R3D-18~\cite{tran2018closer}, DeMamba~\cite{DeMamba}, UNITE~\cite{kundu2025universalsyntheticvideodetector}) tend to underfit with many classes intermingled. Notably, our approach separates multiple classes well.\\

\noindent\textbf{Detection Performance:} For AI-generated video detection, our approach comprehensively surpasses other strong baselines and competitive methods. In the 100-shot scenario, it achieves the highest average detection accuracy of 91.0$\%$, outperforming the state-of-the-art AI-generated video detection method UNITE~\cite{kundu2025universalsyntheticvideodetector} by 8.8$\%$. Though all methods achieve an average detection accuracy of over 70$\%$, it should be emphasized that other methods tend to predict all videos as fake videos, thus achieving extremely low detection accuracy on the real subset (HD-VG and Vript). In contrast, our approach maintains a balance and achieves reasonable performance across all subsets. Even in the 10-shot or 1-shot scenario, our approach consistently outperforms all other mainstream methods. 

\begin{table*}[ht]
\begin{adjustbox}{max width=\linewidth}
\begin{tabular}{c|c|c|cccccccccc}
\toprule
Scheme & Row & Quantization Scale &Real &T2VZ &MS &VC2 &Pika &SVD &MuseV &Mora &CogV & Avg. \\ \midrule
Ours &1 &$s_1=[0.50,1.20,1.90,2.60],s_{2,3}=[0.40,1.00,1.60,2.20]$       &\underline{58.1} &96.7 &76.2 &\textbf{77.5} &\underline{65.7} &78.1 &86.0 &\underline{71.4} &\underline{95.0} &\cellcolor{gray!35}\textbf{78.3} \\ \midrule
\multirow{2}{*}{Channel Disparity} 
&2 &$s_1=[0.50,1.20,1.90,2.60],s_{2,3}=[0.48,1.15,1.85,2.55]$            &56.4 &98.7 &75.4 &74.3 &\textbf{66.2} &\textbf{78.6} &79.8 &68.3 &93.5 &\cellcolor{gray!35}76.8 \\
&3 &$s_1=[0.50,1.20,1.90,2.60],s_{2,3}=[0.20,0.60,1.00,1.40]$            &\textbf{60.1} &\underline{98.8} &75.0 &75.2 &62.8 &73.6 &83.6 &70.1 &82.2 &\cellcolor{gray!35}75.7 \\ \midrule
\multirow{2}{*}{Range Shift} 
&4 &$s_1=[0.20,0.40,0.60,0.80],s_{2,3}=[0.10,0.30,0.50,0.70]$            &55.7 &98.0 &70.3 &71.3 &62.9 &81.3 &\underline{86.2} &\textbf{73.1} &91.1 &\cellcolor{gray!35}76.8 \\
&5 &$s_1=[2.00,3.00,4.00,5.00],s_{2,3}=[1.80,2.70,3.60,4.50]$            &56.5 &96.1 &77.1 &68.6 &65.3 &74.3 &77.8 &62.6 &89.1 &\cellcolor{gray!35}74.2 \\ \midrule
\multirow{2}{*}{Channel Symmetry} 
&6 &$s_1=[0.40,1.00,1.60,2.20],s_{2,3}=[0.40,1.00,1.60,2.20]$            &56.5 &96.7 &73.1 &63.0 &65.5 &77.4 &85.6 &70.4 &92.3 &\cellcolor{gray!35}75.6 \\
&7 &$s_1=[0.50,1.20,1.90,2.60],s_{2,3}=[0.50,1.20,1.90,2.60]$            &55.8 &97.4 &74.5 &70.7 &62.2 &78.0 &85.2 &71.3 &93.0 &\cellcolor{gray!35}76.5 \\ \midrule
\multirow{4}{*}{Scale Cardinality} 
&8 &$s_1=[1.00],s_{2,3}=[1.00]$                                          &58.0 &92.1 &72.7 &68.1 &57.9 &75.2 &84.5 &69.9 &89.1 &\cellcolor{gray!35}74.2 \\
&9 &$s_1=[0.50,1.50],s_{2,3}=[0.40,1.30]$                                &57.3 &97.3 &75.2 &66.8 &61.0 &80.8 &83.2 &65.8 &93.0 &\cellcolor{gray!35}75.6 \\
&10 &$s_1=[0.50,1.50,2.50],s_{2,3}=[0.40,1.30,2.10]$                     &56.7 &97.6 &\underline{78.4} &\underline{75.8} &61.9 &80.8 &\textbf{86.4} &70.2 &\textbf{95.3} &\cellcolor{gray!35}\underline{78.1} \\
&11 &$s_1=[0.50,1.00,1.50,2.00,2.50],s_{2,3}=[0.40,0.80,1.30,1.70,2.10]$ &56.3 &\textbf{99.2} &\textbf{78.7} &75.6 &60.8 &\underline{78.3} &84.2 &70.3 &94.3 &\cellcolor{gray!35}77.5 \\ \bottomrule
\end{tabular}
\end{adjustbox}
\caption{Analysis of quantization scale settings from four perspectives: channel disparity, range shift, channel symmetry, and scale cardinality.}
\label{tab:quantization}
\end{table*}

\subsection{Ablations and Analysis}


\noindent\textbf{Ablation Studies:} We conduct ablation studies on the key components of our approach, including AOCT, MSQR, the temporal stream (Temporal), and the RGB stream (RGB). The results are shown in Tab.~\ref{tab:ablation}. 
(1) \textbf{Effectiveness of AOCT}: Comparing rows 1 and 2, quantization in the adaptive color space significantly improves mAP by 10.9\%, indicating that it helps expose artifacts hidden in RGB frames. 
(2) \textbf{Effectiveness of MSQR}: Comparing rows 2 and 3, multi-scale quantization brings an additional 4.8\% improvement in mAP, demonstrating the benefit of using multiple quantization scales and channel-specific quantization factors. 
(3) \textbf{Effectiveness of the temporal stream}: Comparing rows 3 and 5, adding the temporal stream yields a significant 10.1\% improvement in mAP, highlighting the importance of temporal information for video attribution. 
(4) \textbf{Effectiveness of the RGB stream}: Comparing rows 5 and 6, introducing the RGB stream leads to a modest 1.4\% improvement in mAP, indicating that the original RGB information also provides useful cues for attribution.\\

\noindent\textbf{Analysis of Fusion Weights:} In Sec.~\ref{sec:triple}, we employ a triple-stream feature fusion, and assign different weights to spatial residual stream ($w_{spatial}$), temporal residual stream ($w_{temporal}$) and RGB context stream ($w_{RGB}$). 
(1) \textbf{Selection of ($w_{temporal}$)}: We first set the RGB weight to 0.1, and vary the temporal weights $w_{temporal}$ from 0.1 to 0.6, maintaining the sum of $w_{spatial}$, $w_{temporal}$, and $w_{RGB}$ equal to 1. Results in Tab.~\ref{tab:temporal_weights} show that our approach is robust to different temporal weights $w_{temporal}$, and it achieves the maximum mAP when the temporal weight $w_{temporal}$ is set to 0.3. 
(2) \textbf{Selection of ($w_{RGB}$)}: We subsequently set $w_{temporal}:w_{spatial}$ to 1:2, and vary the RGB weight $w_{RGB}$ from 0.05 to 0.30, maintaining the sum of $w_{spatial}$, $w_{temporal}$ and $w_{RGB}$ as 1. It is evident in Tab.~\ref{tab:rgb_weights} that our approach is also robust to different RGB weights $w_{RGB}$, and the RGB weight $w_{RGB}$ of 0.10 facilitates the approach in achieving the optimal mAP of 78.3$\%$.\\

\begin{figure*}[htbp]
    \centering
    \includegraphics[width=\linewidth]{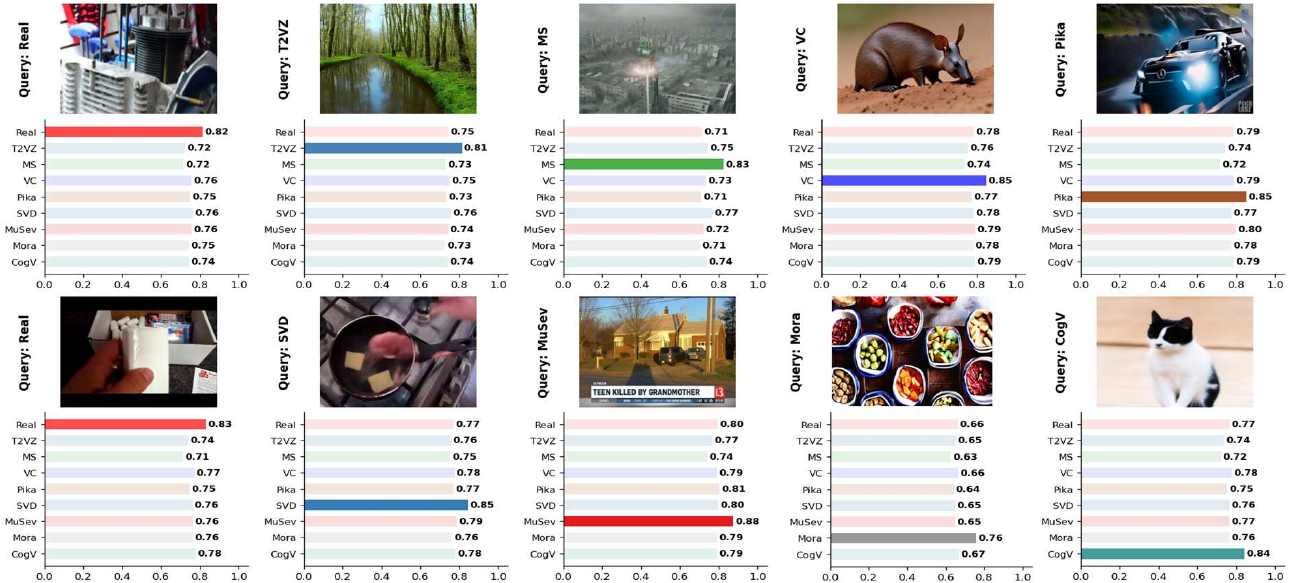}
    \caption{Visualization of video attribution results on hard cases for AI-generated video detection. We select videos misclassified by other detection methods and visualize the attribution results of our approach on these challenging examples. 
    }
    \label{fig:visual}
\end{figure*}

\noindent\textbf{Analysis of the Transformation Matrix:}
Our approach projects video frames into an adaptive color space via a transformation matrix. To evaluate its effectiveness in exposing artifacts, we compare it with several predefined and adaptive alternatives, including the Uniform Difference Matrix (UDM), DCT matrix, YCbCr matrix, YUV matrix, and R/B-channel-dominated adaptive matrices (obtained by replacing the constraint in Eq.~\ref{eq:green_constraint} with $m_{1,R(B)} \ge \tau$). Detailed formulations are provided in the \textit{Supplementary Material}. As shown in Tab.~\ref{tab:matrix}, the UDM and DCT matrix perform the worst, highlighting the importance of a suitable color space. Moreover, adaptive matrices consistently outperform predefined ones, with the G-channel-dominated variant performing best.\\

\noindent\textbf{Analysis of Quantization Scales:} We compare our selection of quantization scales with other choices from four perspectives, and the results of attribution mAP are reported in Tab.~\ref{tab:quantization}. 
(1) \textbf{Channel disparity}: To investigate the channel disparity sensitivity, the difference between $s_1$ and $s_{2,3}$ increases in row (2) and decreases in row (3), and both of them reduce performance. The results indicate that an excessively large difference can lead to excessive information loss, while an excessively small difference can make the method degenerate into symmetric quantization, making it impossible to fully utilize the channel characteristics. Our selection achieves the best balance. 
(2) \textbf{Range shift}: To study the impacts of the range of quantization factors, we narrow down (row 4) or expand (row 5) the range of factors as a whole. Clearly, fine-grained quantization captures minor pixel jitter, while coarse-grained quantization captures structural generation artifacts, and our selection strikes an equilibrium between them. 
(3) \textbf{Channel symmetry}: To verify the effectiveness of asymmetric quantization factors for different channels, we compare asymmetric quantization (row 1) with symmetric quantization (row 6 and row 7), and find that our selection achieves the best performance. This can be ascribed to the different sensitivities of human eyes and generative models to luminance artifacts, and an asymmetric design can more accurately strip away the generation traces of different channels. 
(4) \textbf{Scale cardinality}: When we scale the number of factors from one to five, the mAP increases and then decreases, and it reaches the peak of 78.3$\%$ when we utilize four factors.\\


\noindent\textbf{Analysis of Robustness to Video Degradation:} To assess the robustness against common video processing operations, we apply H.264 compression (CRF=12, 18, and 24) and central crop (maintain 90$\%$, 70$\%$, and 50$\%$ of the whole frame). We evaluate mAP of our approach and other mainstream methods on the video attribution task, which is shown in Tab.~\ref{tab:robustness}. As the degree of video degradation intensifies, the mAP of other methods drops significantly, even below 20$\%$, which indicates that video compression can cause artifacts left by generative models to deform or disappear, and center crop may miss some artifacts. 
Nevertheless, even in the most severe cases, our approach maintains strong robustness, with the mAP remaining above 65$\%$, suggesting that the fingerprints identified by our approach are the essential artifacts that are not easily disturbed.
\\

\noindent\textbf{Analysis of Computational Efficiency:} We compare the time consumption for video attribution between other mainstream methods and our approach. For ESSP~\cite{chen2024single}, LOTA~\cite{Wang_2025_ICCV} and our approach, we compute the time for constructing the 100-shot database as Preparation Time, and compute the time for retrieving 1000 videos as Attribution Time. 
For DeMamba~\cite{DeMamba} and UNITE~\cite{kundu2025universalsyntheticvideodetector}, the time for 100-shot training to convergence and for classifying 1000 videos are regarded as \textit{Preparation Time} and \textit{Attribution Time}, respectively.
Results in Tab.~\ref{tab:efficiency} demonstrate that our approach requires the least total time for attribution, only 1332.30 seconds, outperforming all other methods by more than 685.55 seconds.
\\


\noindent\textbf{Visualization of Attribution in Hard Cases:}
We apply current AI-generated video detection methods to GenVidBench and collect the videos misclassified by these detectors as hard cases. We then apply our approach to these samples and report the feature similarities between query features and database features in Fig.~\ref{fig:visual}. Our approach not only correctly classifies them as real or fake, but also accurately attributes the fake videos to their corresponding generative models.

\section{Conclusion}
We propose a training-free and efficient framework for AI-generated video attribution by formulating attribution as an instance retrieval task. Through adapted orthogonal color transformation, multi-scale quantized residual, and temporal-semantic aggregation, our method effectively exposes subtle generation artifacts and learns robust video representations. 
Extensive experiments on GenVidBench show that our approach consistently outperforms existing baselines in both source attribution (with a Rank-1 accuracy of 20.5$\%$ and a mean Average Precision of 16.6$\%$) and deepfake detection, while maintaining favorable computational efficiency. 
Extensive ablation studies and additional analyses further validate the effectiveness of each proposed module and strong robustness of our design to video degradations.

{
    \small
    \bibliographystyle{ieeenat}
    \bibliography{main}
}


\end{document}